\def\year{2020}\relax
\documentclass[letterpaper]{article} 
\usepackage{aaai20}  
\usepackage{times}  
\usepackage{helvet} 
\usepackage{courier}  
\usepackage[hyphens]{url}  
\usepackage{graphicx} 
\usepackage{graphicx}  
\usepackage{epsfig}
\usepackage{graphicx}
\usepackage{amsmath}
\usepackage{amssymb}
\usepackage{subfig}
\usepackage{multirow}
\usepackage{color}
\usepackage{algorithm}
\usepackage{algorithmic}
\usepackage{hhline}

\begin{document}

\title{Rethinking Temporal Fusion for Video-based Person Re-identification on Semantic and Time Aspect}

\author{
Xinyang Jiang\textsuperscript{\rm 1},
Yifei Gong\textsuperscript{\rm 1},
Xiaowei Guo\textsuperscript{\rm 1},
Qize Yang \textsuperscript{\rm 2}, 
Feiyue Huang\textsuperscript{\rm 1}, \\
\Large \textbf{Weishi Zheng \textsuperscript{\rm 2}, 
Feng Zheng \textsuperscript{\rm 3}, 
Xing Sun\textsuperscript{\rm 1}\thanks{Corresponding Author (winfredsun@tencent.com)}} \\
\textsuperscript{\rm 1} Tencent Youtu Lab, Shanghai, China \\
\textsuperscript{\rm 2} Sun Yat-sen University, Shenzhen, China \\
\textsuperscript{\rm 2} Southern University of Science and Technology, Shenzhen, China \\
}

\maketitle

\begin{abstract}

Recently, the research interest of person re-identification (ReID) has gradually turned to video-based methods, which acquire a person representation by aggregating frame features of an entire video. However, existing video-based ReID methods do not consider the semantic difference brought by the outputs of different network stages, which potentially compromises the information richness of the person features. Furthermore, traditional methods ignore important relationship among frames, which causes information redundancy in fusion along the time axis. To address these issues, we propose a novel general temporal fusion framework to aggregate frame features on both semantic aspect and time aspect. As for the semantic aspect, a multi-stage fusion network is explored to fuse richer frame features at multiple semantic levels, which can effectively reduce the information loss caused by the traditional single-stage fusion. While, for the time axis, the existing intra-frame attention method is improved by adding a novel inter-frame attention module, which effectively reduces the information redundancy in temporal fusion by taking the relationship among frames into consideration. The experimental results show that our approach can effectively improve the video-based re-identification accuracy, achieving the state-of-the-art performance. 
\end{abstract}

\section{Introduction}
\begin{figure}[!ht]

\centering
\includegraphics[width=0.45\textwidth]{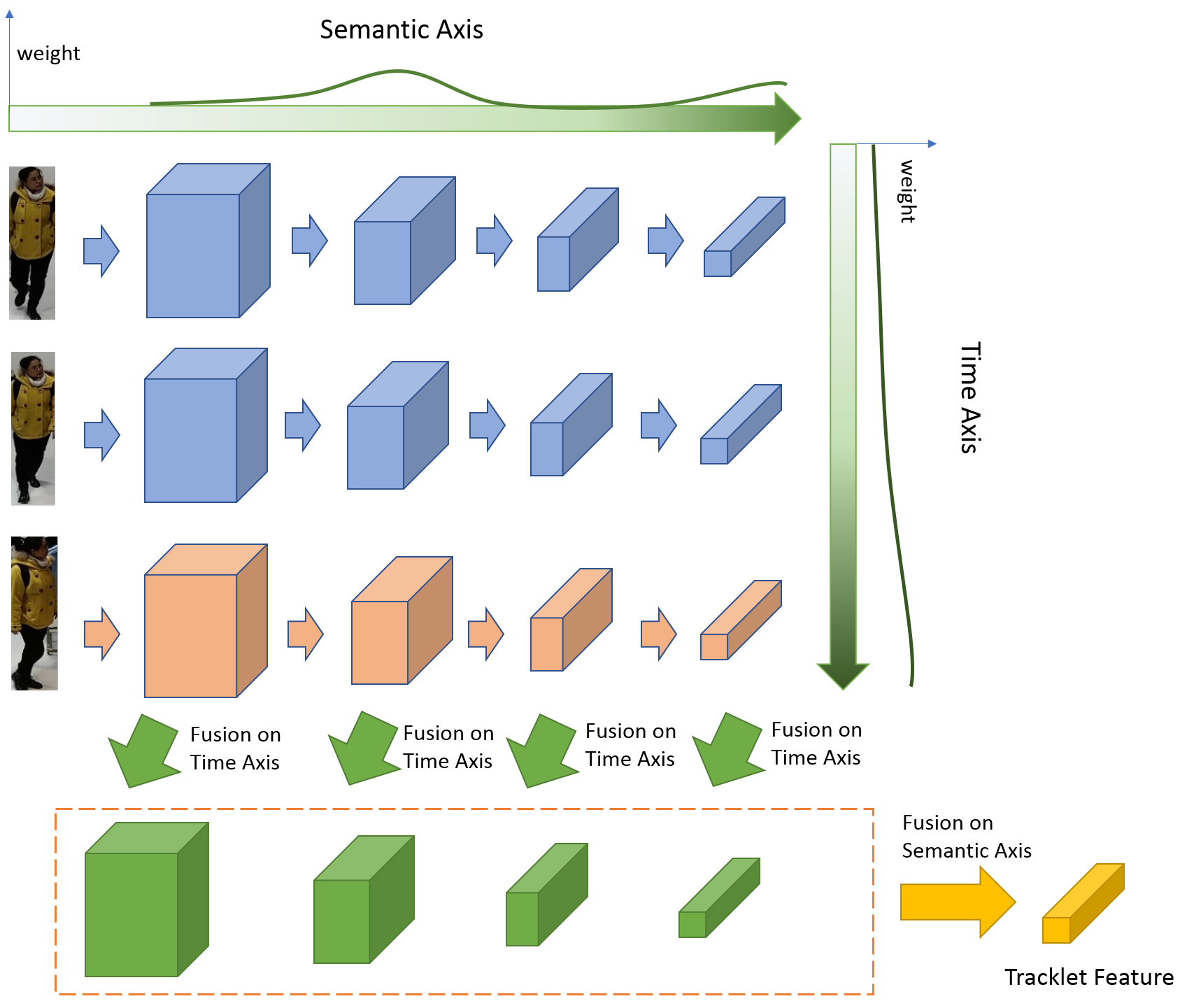} 
\caption{Illustration of temporal fusion. The cuboids represent features of different temporal and semantic levels. The time axis (vertical) represents the image frames in chronological order. The semantic axis (horizontal) represents the frame feature maps extracted from different stages of a CNN, from low-level local feature to high-level semantics. We conduct temporal and semantic fusion to select and weight features from different time and different semantic levels and aggregate the selected features.} 
\label{fig_attention}
\end{figure}

Person re-identification (ReID) is an important technology to match images of pedestrians in different, non-overlapping cameras. During the past few years, 
person re-identification has drawn increasing attention due to its wide applications in surveillance, tracking, smart retail, etc. 
Unlike standard image-based re-identification approaches, video-based re-identification directly takes video/tracklet (i.e. a sequence of images) as input and learns a feature to represent the entire tracklet in an end-to-end fashion, which captures more information from multiple frames in the tracklet, such as temporal cues, variant views, and poses, etc. 

One of the key problems in video-based re-identification is temporal fusion, which is to aggregate feature from each time frame into a comprehensive representation of the tracklet. This paper tries to rethink the temporal fusion problem on the dimension of time and semantics, as well as proposing a unified temporal fusion framework based on temporal and semantic attention. As shown in Figure \ref{fig_attention}, we model the temporal fusion process in a rectangular coordinates system. Frame features are first extracted from different stages of a CNN-based network (the output from different layers in the CNN). The temporal fusion of frame features is performed on the time axis, selectively aggregating these features of different frames, while the semantic fusion aggregates the temporally-fused features that are outputs from different stages of the CNN-based network into a comprehensive tracklet feature. The objective of temporal fusion is to select and weight distinctive frame features, while semantic fusion aggregates tracklet features of different semantic scope. To sum it up, the feature fusion is conducted in two key aspects: time and semantics.


On the semantic axis, frame-feature aggregating at different stages captures semantic information from different levels. Fusing frame information at early stage results in richer temporal information in low-level structural information, while aggregating frame information at late stage results in more information in high-level semantics. 
Thus, a good temporal fusion method should be able to aggregate frame information in multiple semantic levels, i.e. to fuse frame feature-maps at multiple CNN stages. As shown in Figure \ref{fig_attention}, the feature maps of four network stages are fused with different importance weights on the semantic axis.  

Most of the existing video-based ReID methods aggregate frame features in single stages, such as late fusion ReID methods in \cite{zhou2017see,liu2018video} and early fusion methods in video classification and action recognition \cite{simonyan2014two,karpathy2014large}. We propose a novel multi-stage fusion method to fuse feature-maps with multiple semantic levels. Furthermore, a novel semantic attention module is proposed to adaptively assign importance weights of different semantic levels based on the content of the tracklets. 

On the time axis, due to the visual similarity between consecutive frames, a tracklet usually contains a large amount of redundant information, which causes some of the redundant and unimportant frames having large importance weight. A good temporal fusion method should be able to select the important frames on the temporal axis while giving the redundant frames lower weight. As shown in Figure \ref{fig_attention}, the frames of the first two rows have a similar visual appearance, so they are assigned to lower attention weight, while the third frame with a side view is assigned to a higher weight. 



State-of-the-art video-based ReID approach uses attention method to assign different attention weights to different time frames during temporal fusion \cite{chen2018video,li2018diversity}. Most of the existing attention based approaches obtain the frames' attention based on its own content, 
and do not consider the relationships among frames to lower the importance weight for redundant frames and reward frame with distinct features. 
In this paper, we propose an inter-frame attention method that obtains attention weights based on a frame's relationship with others. 

In conclusion, our goal is to design a temporal fusion method for video-based ReID, which has low information loss in semantic aspect and low information redundancy in temporal aspect. To achieve this goal, we propose a multi-stage fusion framework select appropriate frame features along both semantic and time axis. Our contributions are as follows. 1) On semantic aspect, we propose a novel multi-stage fusion method that uses a semantic level attention module to select and fuse appropriate features from all semantic levels. 
2) On temporal aspect, we propose a novel intra/inter-frame attention method, which is the first attention-based fusion method to consider the inter-relationship among frames. 

We verify the effectiveness of our proposed method on three public datasets. The experiments show the multi-stage fusion framework and the intra/inter-frame attention can effectively improve the performance of video-based re-identification.

The rest of the paper is organized as follows. In section 2 we give a brief overview of the existing video-based ReID methods. Then we elaborate our proposed multi-stage fusion framework and the intra/inter-frame attention approach in section 3. In section 4 we report our experiment results. 

\begin{figure*}[!ht]
\centering
\includegraphics[width=0.7\textwidth]{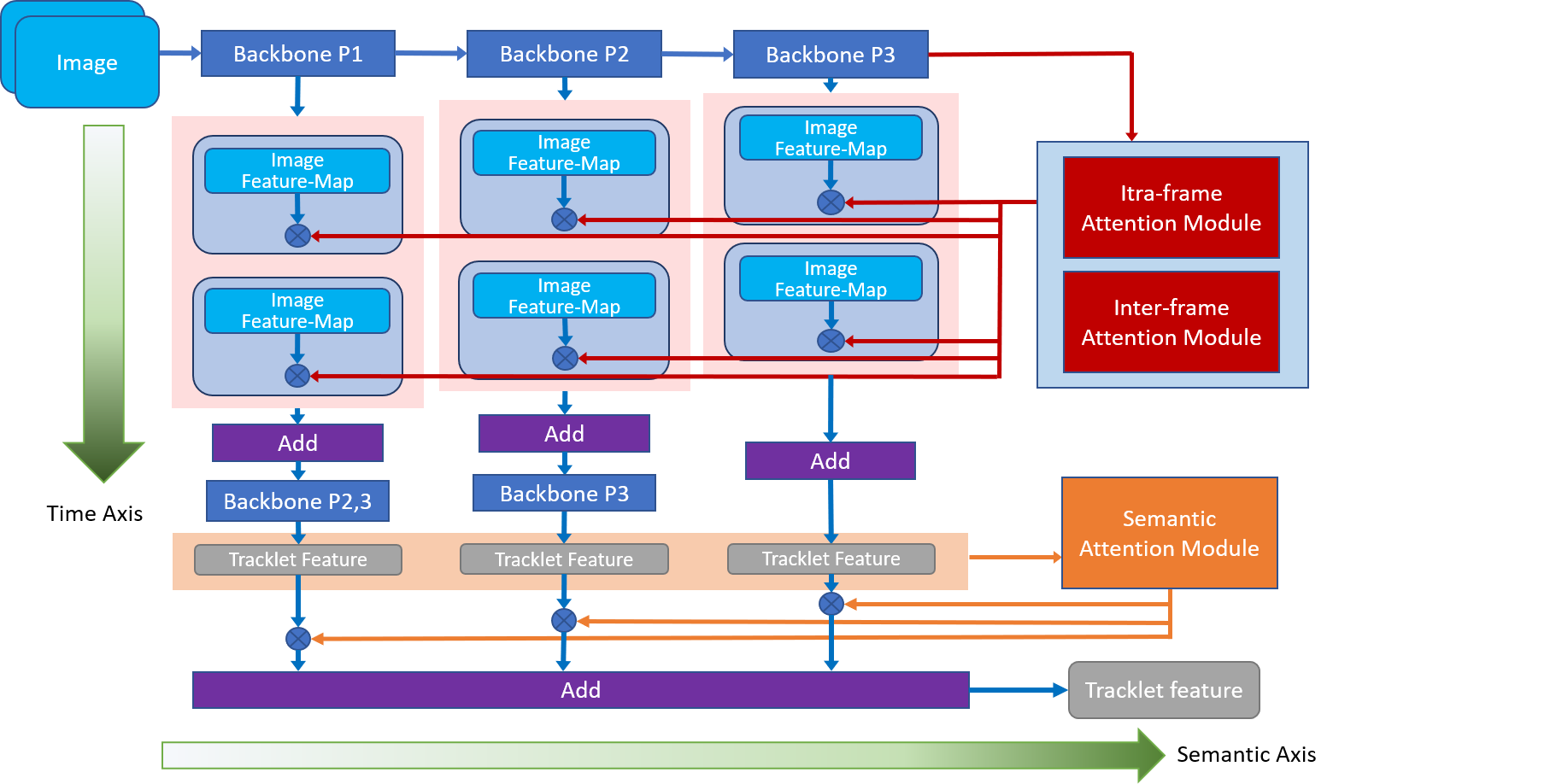} 
\caption{The illustration of Multi-stage Temporal Fusion with intra/inter-frame attention. On the semantic axis, the backbone is broken into three parts. Firstly, on the time axis, each of their outputs are fused using a weighted average with the importance weights from intra/inter-frame attention module. Then the fused feature-maps are encoded into tracklet features at different semantic levels. Then, on semantic axis, these tracklet features are fused using a weighted average with the importance weights from semantic attention module into the final tracklet feature. } 
\label{fig_framework}
\end{figure*}

\section{Related Works}
Derived from multi-camera tracking \cite{huang1997object}, person re-identification as an independent computer vision task was first proposed in \cite{gheissari2006person}. 
Through years of advance in the community cognition towards the topic \cite{zheng2016person} and development in some fundamental techniques, person ReID has grown as a heated research domain with prospective applications in reality. 

\noindent \textbf{Image based person ReID}. Since the problem can be simplified as finding the most similar image in the database given a query image, two components play the key roles during the process---pedestrian description \cite{gheissari2006person}
and distance metric learning \cite{yang2006distance,koestinger2012large} 
. As the CNN-based models prevail, two types of models are commonly deployed. The first type treats the problem as image classification \cite{ahmed2015improved} while the second takes a Siamese model and use image pairs \cite{radenovic2016cnn} or triplets \cite{schroff2015facenet,hermans2017defense} as input. More recently, explicitly leveraging image local stripes \cite{sun2018beyond}
or implicitly use attention scheme show decent results \cite{wang2018mancs}\cite{si2018dual} 
in many dataset. 

\noindent \textbf{Temporal attention for video-based person ReID}. Different from image-based task, video-based analysis \cite{karpathy2014large,simonyan2014two}
takes in a sequence of images thus can leverage the extra temporal information. In video-based person ReID, attention models highlight informative frames by assigning them higher scores. Zhou et al.\cite{zhou2017see} combine spatial and temporal information to jointly learn features and metrics. Liu et al.\cite{liu2017hydraplus} propose a multi-directional attention module to exploit the global and local contents for image-based person ReID. Researchers like Li et al. \cite{li2018diversity} and Fu et all. \cite{fu2019sta} propose using Spatial-temporal attention model that automatically discovers a diverse set of distinctive body parts. \cite{hou2019vrstc} proposes a method that computes attention weights based on patches in adjacent frames, which considers the relationship between neighboring frames, while our paper considers the relationship among all frames and uses Relation Network to mine complex relationship feature instead of simple predefined feature similarity. In this paper, we propose using attention methods on both time (frame-level) axis and semantic axis (model stages) and design a novel intra/inter attention module. 

\noindent \textbf{Feature aggregation for video-based person ReID}. Fusing features in an early or late stage with average or max pooling to enhance the feature expression ability is widely used in video analysis \cite{karpathy2014large,snoek2005early}. Recurrent Neural Networks have also been used to integrates the surrounding information \cite{mclaughlin2016recurrent,zhou2017see,xu2017jointly}. K L et al.\cite{murthy2018deep} performs operator-in-the-loop feature fusion from multiple camera images for person re-identification. Johnson et al. \cite{johnson2018person} fuse the handcrafted feature and deep feature to complement the global body features. In this work, we propose a novel multi-stage fusion framework that integrates features from multiple stages into a  two-branch structure.

\section{Methods}

We propose a new video-based person re-identification model with a novel temporal fusion method that has advantages in both temporal and semantic aspect. To reduce the information redundancy, we propose an intra/inter-frame attention model. To reduce the information loss in the semantic aspect, we propose a multi-stage fusion structure. 

We first briefly introduce the general structure of the proposed method. As shown in Figure \ref{fig_framework}, the input of a video-based person re-identification model is an image sequence containing a certain person, called a tracklet. Our framework contains multiple branches, each performs image-level fusion on time axis on different network stages, from early to late. Then, on semantic axis, the fused tracklet features of all the branches are fused together to get the final feature of the tracklet. We use a novel inter/intra relational attention module to get the importance weight for fusion on time axis, and use a semantic attention module to get the importance weight for fusion on semantic axis. 

\begin{figure*}[!ht]
\centering
\includegraphics[width=1\textwidth]{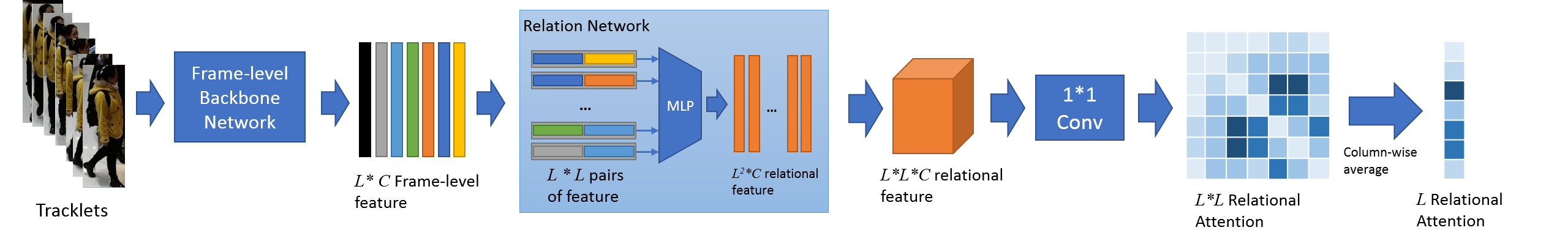} 
\caption{Inter-Frame Attention Module with Relation Network} 
\label{fig_relation_attention}
\end{figure*}

\subsection{Time Aspect: Intra/Inter-Frame Attention}

On the time axis, we need to conduct an image-level fusion to merge the features of all the images in a tracklet into one tracklet feature. In order to filter redundant frames and emphasize on important frames, we propose a novel intra/inter-frame attention method. 

To fuse image features in a person tracklet, many existing methods adopt average or max pooling across the frames. However, as the quality and content vary drastically across frames, it is essential to weaken the impact of noisy, low-quality frames and strengthen the impact of high-quality informative frames. Thus, each frame should be assigned to an important weight in temporal fusion. We call this kind of fusion method attention-based method. Given a tracklet with $L$ frames, to get a fused tracklet feature denoted as $\mathbf{f}_{fused}$, attention-based approaches adopt a weighted average pooling operation to fuse the image features: 
\begin{equation}
\mathbf{f}_{fused} = \frac{1}{N}\sum_i^N(a_i\mathbf{f}_i)
\label{eq_time_level_fusion}
\end{equation}

where in this paper, the attention weight $a_i$ is the average of intra-frame attention $w_i$ and inter-frame attention $v_i$:
\begin{equation}
    a_i = (w_i + v_i) / 2
    \label{eq_inter_intra_attention}
\end{equation}

In the following subsections, we elaborate the proposed intra/inter-frame attention method and how intra-frame and inter-frame attention is computed. 

\subsubsection{Intra-frame Attention}
Most of the existing attention-based approaches obtain the attention of a frame based on its own quality and content (e.g., resolution, occlusion, camera angle, etc). We call this type of attention-based approaches intra-frame attention. 

Our implementation of intra-frame attention is as follows. Given a video tracklet containing $L$ frames, we first use a frame-level backbone network to extract feature for each frame $i$, denoted as $\mathbf{f_i}$. Then, a binary regressor $A_s$ is used to predict an importance score $w_i$ for each frame:
\begin{equation}
    w_i = A_s(\mathbf{f}_i)
\label{eq_intra_frame_attention}
\end{equation}

\subsubsection{Inter-frame Attention}
To focus more on frames with distinct features and reduce redundancy, frames with similar visual appearance should be assigned to lower attention weights while visually distinct frames should be assigned to higher attention weights. 
As a result, the importance of a frame not only depends on its own content, but also its relationship and differences with the other frames in the tracklet.  We call this kind of relationship-based attention inter-frame attention. 

Figure \ref{fig_relation_attention} is an illustration of our proposed relation network based on inter-frame attention module. Same as the intra-frame attention, we use the same frame-level backbone network to extract frame-level feature $f_i$ for each frame in the tracklet. The most straight forward way to obtain the correlation between two frames is to use a similarity measure, such as euclidean distance or cosine distance. Thus, for any frame feature in a tracklet with length $L$, denoted as $\mathbf{f}_i$, its inter-frame attention is the mean distance between $\mathbf{f}_i$  and every other frame $\mathbf{f}_j$ in the tracklet:
\begin{equation}
    v_{ij} = \frac{1}{L}\sum_{j} d(\mathbf{f}_i, \mathbf{f}_j)
\label{eq_euclidean_attention}
\end{equation}
where $d$ is a predefined similarity and the euclidean distance is used in our experiments. 


Furthermore, we argue that besides simple similarity, much richer correlation information is needed for inter-frame attention. Thus, we propose to apply a relation network to obtain a relation embedding for each pair of frames in the tracklet. Given a video tracklet containing $L$ frame features. The inputs of the relation network are generated by concatenating each pair of the $L$ $d_f$-dimensional frame features. Then, $L*L$ vectors with $2*d_f$ dimensions are embedded into a $d_r$ dimensional relation space by a multi-layer perception. As a result, the relation network (denoted as function $P$) generate a pair-wise relation embedding $\mathbf{r}_{i, j}$ for each pair of the frame feature $\mathbf{f}_i$, $\mathbf{f}_j$, as follow:
\begin{equation}
    \mathbf{r}_{i, j} = P([\mathbf{f}_i, \mathbf{f}_j]) + P([\mathbf{f}_j, \mathbf{f}_i]). 
\end{equation}
Noted that the relational attention features from both directions are added together in order to obtain a symmetric attention matrix. 

The relational embedding is then reshaped into a $L * L * d_r$ tensor. We feed the tensor into a $1 * 1$ convolutional layer with $1$ output channel and obtain a $L * L$ attention matrix $A$. Each element $a_{i, j}$ in the matrix indicate the attention weight of the $i$th frame based on its relationship with the $j$th frame. The inter-frame attention of the $i$th frame $v_i$ is the mean value of all $L - 1$ attention weights respect to the other frames, which is equivalent to apply a column-wise average operation on the attention matrix:
\begin{equation}
    v_i = \frac{1}{L - 1}\sum_{j=0,j\neq i}^{L-1}ReLU(\theta\mathbf{r}_{i,j})
    \label{eq_rn_attention}
\end{equation}
where $\theta$ is the parameter of the $1*1$ convolution layer.


\subsection{Semantic Aspect: Multi-Stage}

\begin{figure}[!ht]
    \centering
    \subfloat[Late Fusion]{
    \begin{minipage}[t]{0.14\textwidth}
        \centering
        \includegraphics[height=2.1\textwidth]{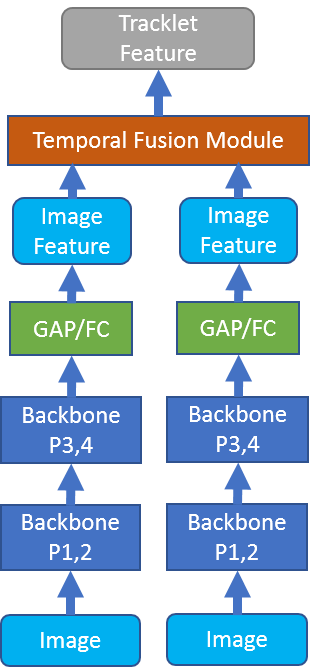}
    \end{minipage}
    }
    \subfloat[Early Fusion]{
    \begin{minipage}[t]{0.14\textwidth}
        \centering
        \includegraphics[height=2.1\textwidth]{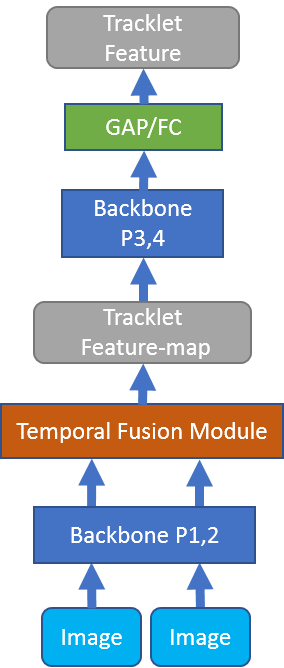}
    \end{minipage}
    }
    \subfloat[Multi-stage Fusion]{
    \begin{minipage}[t]{0.14\textwidth}
        \centering
        \includegraphics[height=2.1\textwidth]{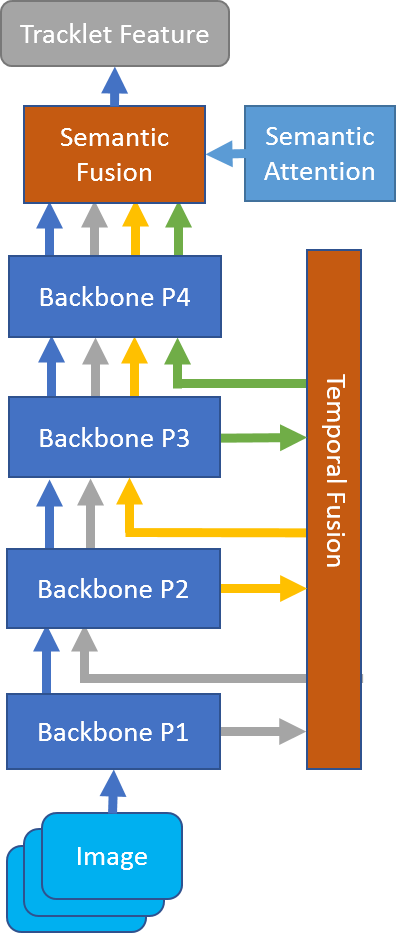}
    \end{minipage}
    }
\caption{The illustration of different fusion methods.}
\label{fig_fusion_methods}
\end{figure}

\subsubsection{Multi-stage Fusion}
To reduce the information loss caused by aggregating image features in a single semantic level, we propose to conduct the aforementioned intra/inter-frame fusion on multiple semantic levels. As a result,  the fused tracklet feature vectors from multiple semantic levels should then be fused one more time on semantic level. 

Traditional temporal fusion method only fuses image-level features on one certain stage of the backbone network. As shown in Figure~\ref{fig_fusion_methods} (a), the late-stage fusion fuses the image features at the bottom of the network. It makes sure the information within the frame is fully analyzed by a large amount of network layers, but information among frames in the video track are not explored enough. On the other hand, early-stage fusion fuses the image feature-maps at the earlier stage of the network and uses more layers to analyze the temporal information in the whole tracklet at the cost of insufficient frame level information extraction, as shown in Figure~\ref{fig_fusion_methods} (b). 

Instead of leveraging between late fusion and early fusion, we propose a fusion framework taking advantages of the both, namely the Multi-Stage Fusion. 
The temporal fusion on feature maps is conducted at multiple stages, from early to late.  For example, in Figure \ref{fig_fusion_methods} (c), there are four fusion branches, which fuses the output feature-maps of layers at different stages of a backbone network, from early to late. In this paper, we use the same backbone network to encode the tracklet feature-maps into tracklet feature vectors, i.e. all four branches feed the fused tracklet feature-map back into the subsequent layers in the original backbone network to get the final tracklet feature vectors. 

\subsubsection{Semantic Attention Module}
In this sub-section, we elaborate on our novel attention-based semantic fusion mechanism. Same as the fusion on time axis, we believe that the importance of different semantic levels heavily depends on the content of the tracklet. As a result, we propose a novel semantic attention module to assign different attention weights to  feature vectors on different semantic levels. 

Our implementation of the semantic attention is as follows. Given $K$ video tracklet features from different semantic levels, for each of the branch output features denoted as $g$, we use a softmax classifier $B$ to predict an importance score for every branch, so for the $i$-th branch feature, the importance weight of the $j$-th branch (denoted as $u_{ij}$) is computed as follows:
\begin{equation}
    u_{ij} = B(\mathbf{g}_i)_j
\label{eq_semantic_attention1}
\end{equation}
As a result, the over importance of the $j$-th branch is:
\begin{equation}
    u_j = \frac{1}{K}\sum_{i}B(\mathbf{g}_i)_j
\label{eq_semantic_attention2}
\end{equation}

The final tracklet feature of a tracklet $\mathbf{g}_{fused}$ is computed as follow, 
\begin{equation}
\mathbf{g}_{fused} = \frac{1}{N}\sum_j^N(u_j\mathbf{g}_j)
\label{eq_semantic_level_fusion}
\end{equation}
where $N$ is the number of semantic branches. 
The pipeline of our method is shown in Algorithm 1.

o\begin{algorithm}
\caption{Multi-stage Fusion with Intra/Inter Frame Attention}
\label{alg_1}
\begin{algorithmic}
\REQUIRE
Video tracklet containing $n$ frames
\ENSURE
Tracklet feature $\mathbf{g}_{fused}$
\FOR{each frame in tracklet with index $i$}
\STATE Extract frame level feature $\mathbf{f}_i$
\STATE Compute intra-attention $w_i$ with Eq.\ref{eq_intra_frame_attention}
\STATE Compute inter-attention $v_i$ with Eq.\ref{eq_rn_attention}
\STATE Compute image-level attention with Eq.\ref{eq_inter_intra_attention}
\ENDFOR
\STATE Normalize image-level attention
\FOR{each network stage with index $i$}
\STATE Obtain the output from the layer at network stage $i$
\STATE Compute tracklet-level feature-map with Eq.\ref{eq_time_level_fusion} 
\STATE Obtain tracklet feature by feeding feature-map back into the subsequent layers of the network
\STATE Obtain semantic attention with Eq.\ref{eq_semantic_attention1} and Eq.\ref{eq_semantic_attention2}
\ENDFOR
\STATE Obtain final output with Eq.\ref{eq_semantic_level_fusion}

\end{algorithmic}
\end{algorithm}

\section{Experiments}
\subsection{Datasets and Evaluation Protocol}
We evaluate the proposed algorithm on three benchmark datasets: PRID2011 \cite{hirzer2011person},  iLIDS-VID \cite{li2018unsupervised}\cite{ma2017person} and MARS \cite{zheng2016mars}. PRID2011 consists of 749 people from two camera
views, 178 of which appear in both cameras. iLIDS-VIDS consists of 300 persons' 600 tracklets, where each person has two tracklets from different cameras with a length of 23 to 192 frames. The MARS dataset contains 1261 persons' over 20,000 tracklets. Each person appears in at least two cameras and has an average of 13.2 tracklets. 

For PRID2011 and iLIDS-VID dataset, following \cite{wang2014person}, the dataset is randomly split into training and testing set for 10 times, each containing $50\%$ of the data. For PRID2011 only the identities appeared in both camera are used. The averaged accuracy is computed over the 10 different training/testing splits. For MARS dataset, we use the training/testing split in \cite{zheng2016mars}, where 631 identities are used for training and the rest are used for testing.  
The re-identification performance is measured by mean average precision (MAP) and rank-n accuracy. 

\subsection{Implementation details}
We used a pre-trained ResNet-50 image-based person re-identification network as the backbone network. The image-based ReID model was first trained on image-based ReID training set including CUHK03 \cite{li2014deepreid}, DukeMTMC \cite{ristani2016MTMC}, Market1501. Then the model was fine-tuned independently on PRID2011, iLIDS-VIDS, and MARS. After that, we used the trained image-based network as the backbone of the multi-stage fusion framework with attention modules and continued fine-tuning the model in an end-to-end video-based way. 
On semantic axis we fuse tracklet features from four stages. The input image was resized to a size of 256 * 128. We used the stochastic gradient descent algorithm to update the weights and used a staircase schedule strategy where the learning rate decayed 0.8 every 20 epochs.  
For each tracklet, the model output a 768-dimensional feature vector.

\subsection{Ablation Study}
\begin{table}
\centering
\small
\caption{Performance comparisons of different fusion methods on MARS.}
\begin{tabular}{|c|c|c|c|c|}
    \hline
    Methods & MAP & top1 & top5 \\
    \hline
    Feature Average & 70.3  & 78.1 & 91.4 \\
    \hline
    Early Fusion & 74.9 & 82.2 & 93.6 \\
    \hline
    Late Fusion & 75.4 & 82.5 & 93.5 \\
    \hline
    MS Fusion (Average) & 75.6 & 83.3 & 94.1 \\
    \hline
    MS Fusion (Semantic Attention) & $\mathbf{77.7}$ & $\mathbf{83.6}$ & $\mathbf{94.0}$ \\
    \hline
\end{tabular}
\label{table_fusion}
\vspace{-0.3cm}
\end{table}

\begin{table}
\small
\centering
\caption{Performance comparisons of different attention methods on MARS.}
\begin{tabular}{|c|c|c|c|}
    \hline
    Attention Methods &  MAP & top1  & top5 \\ 
    \hline
    Average Pooling (late-fusion) & 75.4 & 82.5  & 93.5 \\ 
    \hhline{|====|}
    Intra-frame & 76.6 & 85.3  & 94.4 \\ 
    \hline
    Inter-frame (Euclidean) & 75.7 & 83.9 & 94.1 \\ 
    \hline
    Inter-frame (RN)  & 76.8 & 84.2  & 94.2 \\ 
    \hhline{|====|}
    
    Intra/Inter-frame (Euclidean) & 79.1 & 84.0 & 94.6 \\ 
    \hline
    Intra/Inter-frame (RN) & 82.4 & 85.8 & 95.7\\ 
    \hline
    Intra/Inter-frame (RN) + multi-stage & 85.2 & 87.1 & 96.8 \\ 
    \hline
\end{tabular}
\label{table_attention}
\vspace{-0.3cm}
\end{table}

In this sub-section, we verify the effectiveness of our proposed fusion and attention components by ablation study. 

To verify the effectiveness of multi-stage fusion, we compare the performance of different fusion methods. Table \ref{table_fusion} reports the performances of different fusion approaches on MARS. Following approaches are compared:
\begin{itemize}
    \vspace{-4pt}
    \item{\textbf{Feature Average}}. Training a image-based model and uses the average of the image features as the tracklet feature without any end-to-end video-based training. 
    \vspace{-4pt}
    \item{\textbf{Early Fusion}}. Training an end-to-end video-based model with an early fusion after the 2nd res-block. 
    \vspace{-4pt}
    \item{\textbf{Late Fusion}}. Training an end-to-end video-based model with late fusion after the 4th res-block. 
    \vspace{-4pt}
    \item{\textbf{Multi-Stage (MS) Fusion (Average)}}. Tracklet features from multiple semantic stages are fused with average pooling .
    \vspace{-4pt}
    \item{\textbf{Multi-Stage (MS) Fusion with Semantic Attention}}. Tracklet features from multiple semantic stages are fused with semantic attention. 
\end{itemize}

\begin{table*}
\small
\centering
\caption{The comparisons of our method to the state-of-the-art methods on PRID2011, iLIDS-VID, and MARS datasets.}
\begin{tabular}{|c|c|c|c|c|c|c|}
    \hline
    \multirow{2}{*}{Methods} & \multirow{2}{*}{Fusion Type} & \multirow{2}{*}{Attention Type}  & PRID & i-LIDS-VID & \multicolumn{2}{c|}{MARS} \\ \cline{4-7} 
    & & & top1 & top1 & MAP & top1 \\
    \hline
    See-Forest \cite{zhou2017see} & Late & NA  &79.4 &55.2 & 50.7 & 70.6 \\
    \hline
    AMOC+epicFlow \cite{liu2018video} & Late & NA  & 83.7 & 68.7 & 52.9 & 68.3 \\
    \hline
    Spatial-temporal  \cite{li2018diversity} & Late & Intra-Frame & 93.2  & 80.2 & 65.8 & 82.3 \\
    \hline
    LSTM \cite{gao2018revisiting} & Late & NA & - & - & 73.9 &  81.6 \\
    \hline
    Non-local \cite{liao2018video} & Multistage & NA & 91.2 & 81.3 & 77.0 & 84.3 \\
    \hline
    STA \cite{fu2019sta} &  Late & Intra-Frame & - & - & 80.4 & 85.5 \\
    \hline
    Attribute Driven \cite{zhao2019attribute} &  Late & Intra-Frame Attention  & 93.9 & 86.3 & 78.2 & 87.0 \\
    \hline
    VRSTC \cite{hou2019vrstc} & Late & Inter-Frame & - & 83.4 & 82.3 & $\mathbf{88.5}$ \\
    \hline
    Ours & Multistage & Intra-Inter-Frame / Semantic & $\mathbf{95.8}$ & $\mathbf{87.7}$ & $\mathbf{85.2}$ & 87.1 \\  
    \hline 
\end{tabular}
\label{table_performance}
\vspace{-0.5cm}
\end{table*}


\begin{figure}
\centering
\small
\includegraphics[width=0.5\textwidth]{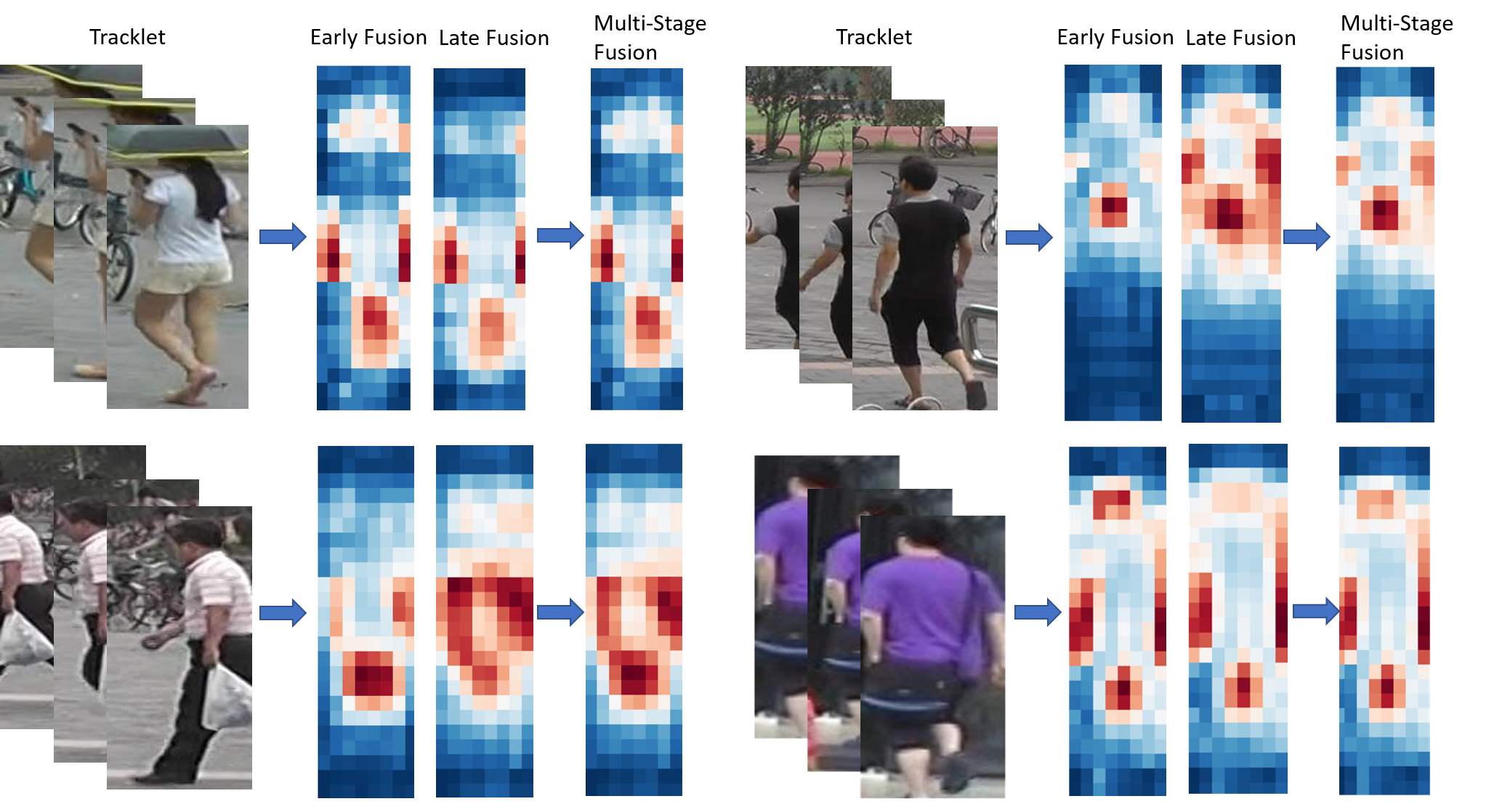} 
\caption{Feature-maps computed by early fusion, late fusion and multi-stage fusion. } 
\label{fig_vis_fusion}
\end{figure}

\begin{figure}
\centering
\small
\includegraphics[width=0.45\textwidth]{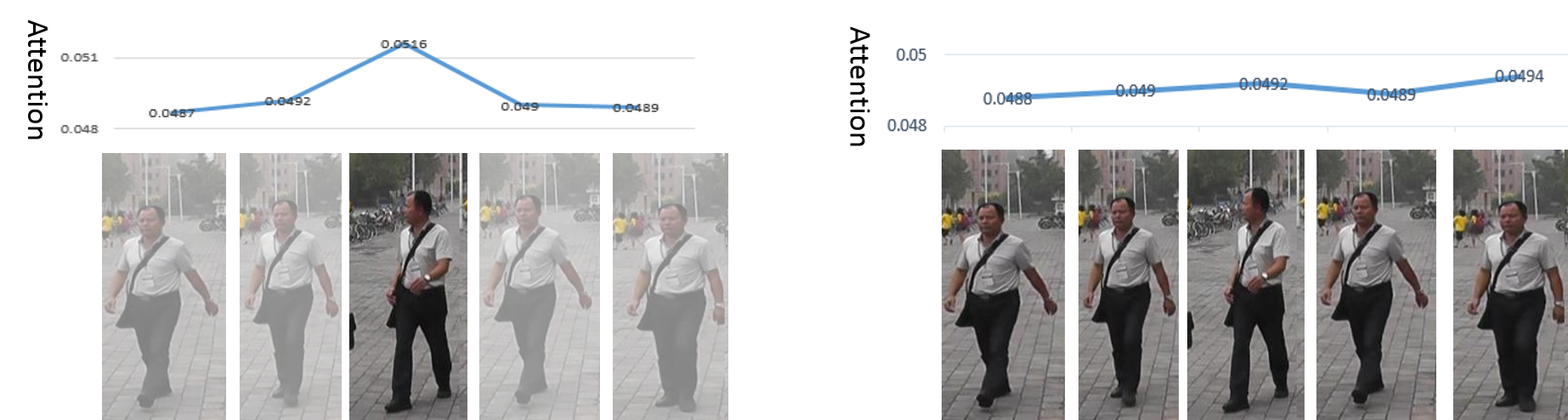} 
\caption{Example of intra/inter-frame attention (left) and inter-frame attention (right) for same tracklet. } 
\label{fig_vis_attention}
\end{figure}


From Table \ref{table_fusion}, we can make following observation. 1) Compared to the image-based model, all end-to-end temporal fusion methods achieve higher performance, which verifies the advantage of adopting an end-to-end video-based approach. 2) Multi-stage fusion methods outperform the single-stage early fusion and late fusion method, showing that taking advantage of multi-level semantics from both early and late stage boosts the video reid performance. 3) Adding semantic attention improves the multi-stage fusion by 2 percent, proving the superiority of adaptively assigning the importance weights to different stages.    

To verify the effectiveness of the intra/inter attention module, we compare the performance different attention methods. Table \ref{table_attention} shows the performance comparison of different attention based methods. We compare following attention approaches:
\begin{itemize}
    \vspace{-4pt}
    \item \textbf{Average Pooling}. A late-fusion model that uses image features with average pooling. Equal attention weights are assigned for each frame. 
    \vspace{-4pt}
    \item \textbf{Intra-frame Attention}. Late-fusion module with  intra-frame attention module. 
    \vspace{-4pt}
    \item \textbf{Inter-frame Attention (Euclidean)}. The euclidean based inter-frame attention module (Eq.(\ref{eq_euclidean_attention})) is integrated into the late fusion baseline. 
    \vspace{-4pt}
    \item \textbf{Inter-frame Attention (RN)}. A Relation Network based attention module  ( Eq.(\ref{eq_rn_attention})) is added to the baseline. 
    \vspace{-4pt}
    
    \item \textbf{Intra/Inter-frame Attention (Euclidean)}. We add euclidean distance based inter-frame attention module to the intra-frame attention baseline.
     
     \vspace{-4pt}  
     \item \textbf{Intra/Inter-frame Attention (RN)}. We add a RN based inter-frame attention module to the intra-frame attention baseline. 
    \vspace{-4pt}    
    \item \textbf{Intra/inter-frame (RN) + Multi-Stage}. Image level fusion with intra/inter-frame attention module are applied on multiple stages and then the multi-level semantic fusion is applied with semantic attention module. 
\end{itemize}

\textbf{Pooling vs. Attention}. From Table \ref{table_attention}, we observe that the attention based approaches perform better than directly average pooling, which verify the effectiveness of adding attention weights during fusion. 


\textbf{Different Attention Approaches}. From Table \ref{table_attention}, we observe that adding an extra inter-frame attention module outperforms intra-frame attention module by 2 to 3 percentage point in terms of MAP, which verifies that obtaining attention based on relationship among frames effectively boosts the performance of video-based ReID. Furthermore, only using inter-frame attention module cannot achieve as high performance as using both inter and intra-frame attention. These results indicate that although the relation network in the inter-frame attention can discover the relationship among frames, its ability to discover the information with a single frame is limited. Therefore, an intra/inter-frame two-branch structure is necessary to make up for its disadvantage.  

\textbf{RN vs. Predefined Correlation}.  Compared to predefined cross correlation (i.e. euclidean distance in Table \ref{table_attention}) , we observe that applying Relation Network in inter-frame attention achieves better performance.  This is because RN is able to mine more complex relation among frames by using deep neural structures other than simple similarity between feature vectors, which is essential for inter-frame attention. 

\textbf{Time axis vs. Semantic Axis}. After adding multi-stage fusion to the intra/inter-frame attention based fusion method, we achieve the best performance,  (i.e. Intra/inter-frame (RN) + multi-stage in table \ref{table_attention}), which verifies the effectiveness of conducting feature selection and fusion on both time and semantic axis.


To demonstrate the advantage of multi-stage fusion, we visualize the feature maps from different semantic levels as shown in Figure \ref{fig_vis_fusion}. Here we can see that feature maps fused at the earlier stage tend to be sparser and focus more on distinct structural information, while feature maps from the later stage tend to have a strong response on a wider range of structures. Compared to only using the feature from the late stage, fusing it with the feature map from an earlier stage enables the model to represent the tracklet more comprehensively. Take the first case with the woman in white as an example, while the late fusion feature map does not have a strong response on the umbrella, a distinctive feature of the tracklet, the feature map from the early stage captures the structural information of the umbrella and helps the fused feature map to better represent the tracklet.

Figure \ref{fig_vis_attention} shows an example of a tracklet and its attention weights computed by the inter-frame attention module and intra-frame attention module. We observe that, the inter-frame attention module assigns a higher weight to the third frame of the tracklet because it is the only side view frame in the tracklet and is more distinct. On the other hand  intra-frame attention assigns similar weights to every frame. This example shows that inter-frame attention has ability to lower the importance of redundant frames and increase the weight of frames with distint visual features.

\vspace{-0.2cm}
\subsection{Comparison with the State-of-the-Art Method}
In table \ref{table_performance}, we compare our approaches with state-of-the-art temporal fusion methods for video ReID approaches. All the comparing approaches use only video tracklets as the inputs, and no re-ranking and multi-query strategy are used in post-processing.  

As shown in table \ref{table_performance}, we compare our method with state-of-the-art approaches that adopt various ways of feature fusion and attention generation. We compare our method with SOTA late fusion methods including See-Forest \cite{zhou2017see}, AMOC+epicFlow \cite{liu2018video} and  \cite{gao2018revisiting} that uses LSTM to fuse image features. We compare our method with SOTA attention based methods (\cite{fu2018sta} \cite{liu2015spatio}, and STA \cite{fu2019sta}) which uses both temporal and spatial attention. An attribute driven methods \cite{zhao2019attribute} that include extra body attribute to boost the acurracy of attention prediction.  Non-local \cite{liao2018video} do not use late fusion, but adopts a 3D CNN network with a non-local attention module. Table \ref{table_performance} shows that our approach outperforms the state-of-the-art approaches on PRID2011, i-LIDS-VID, and achieves highest MAP score on MARS. Note that, on MARS dataset, VRSTC achieves higher top-1 by 1 percentage point compared our method, probably because it also considers inter-frame relationship between the adjacent frames and  uses extra spatial attention to obtain attention weights on local patches. Although not apply spatial attention, Our method outperforms VRSTC by 2.9 percentage points in terms of MAP because 1) our inter-frame attention not only considers relationship between adjacent frames but also all possible frame pairs, and 2) we uses a Relation Network to discover the complex relationship features instead of predefined similarity measure.  

\vspace{-0.2cm}
\section{Conclusions}
This paper proposes a novel temporal fusion method for video-based re-identification. We propose a general temporal fusion method to automatically select features on both semantic and time aspects. On the semantic aspect, our method aggregates feature maps on multiple semantics levels by a multi-branch structure. On the time aspect, we propose an intra/inter-frame attention module to take the relationship between frames into consideration. The experiments verify the effectiveness of our two novel model components and our approach achieves state-of-the-art performance on video-based re-identification benchmarks. 

{\small
\bibliographystyle{aaai}
\bibliography{references}
}

\end{document}